\documentclass{article}
\usepackage{ijcai24}

\usepackage{times}
\usepackage{soul}
\usepackage[hidelinks]{hyperref}
\usepackage{algorithm}
\usepackage{algorithmic}
\usepackage[switch]{lineno}
\usepackage[utf8]{inputenc} 
\usepackage{url}            
\usepackage{booktabs}       
\usepackage{amsfonts}       
\usepackage{amsmath, amsthm}        
\usepackage{nicefrac}       
\usepackage{microtype}      
\usepackage{xcolor}         
\usepackage{graphicx}       
\usepackage{subcaption}     
\usepackage{tabularx, tabulary}
\usepackage{tikz}           
\usepackage{scalerel}       
\usepackage{siunitx}        
\usepackage{float}          
\usepackage[font=small]{caption}        
\usepackage{wrapfig}        
\usepackage{array}
\usepackage{booktabs}
\usepackage{enumitem}       

\definecolor{backcolour}{rgb}{0.95,0.95,0.92}

\usetikzlibrary{positioning}

\newcolumntype{b}{X}
\newcolumntype{m}{>{\hsize=.7\hsize}X}
\newcolumntype{s}{>{\hsize=.3\hsize}X}

\definecolor{mydarkblue}{rgb}{0,0.08,0.45}

\hypersetup{ %
    pdfborder=0 0 0,
    pdfpagemode=UseNone,
    colorlinks=true,
    linkcolor=mydarkblue,
    citecolor=mydarkblue,
    filecolor=mydarkblue,
    urlcolor=mydarkblue,
}

\newcommand{\reals}{{\mathbb R}}

\title{The Extrapolation Power of Implicit Models}

\author{%
Juliette Decugis$^1$ 
\and
\textbf{Alicia Y. Tsai}$^{1, 2}$
\and \\
Max Emerling$^1$ 
\and
Ashwin Ganesh$^1$
\and
\textbf{Laurent El Ghaoui}$^{1, 2}$ \\
\affiliations
$^1$UC Berkeley
$^2$VinUniversity \\
\emails
\{jdecugis, aliciatsai, memerling, aganesh, elghaoui\}@berkeley.edu \\
\{alicia.t, laurent.eg\}@vinuni.edu.vn \\
}

\begin{document}

\maketitle
\begin{abstract}
In this paper, we investigate the extrapolation capabilities of implicit deep learning models in handling unobserved data, where traditional deep neural networks may falter. Implicit models, distinguished by their adaptability in layer depth and incorporation of feedback within their computational graph, are put to the test across various extrapolation scenarios: out-of-distribution, geographical, and temporal shifts. Our experiments consistently demonstrate significant performance advantage with implicit models. Unlike their non-implicit counterparts, which often rely on meticulous architectural design for each task, implicit models demonstrate the ability to learn complex model structures without the need for task-specific design, highlighting their robustness in handling unseen data.
\end{abstract}

\section{Introduction}

Learning to extrapolate, which involves estimating unknown values beyond observed data, is a fundamental aspect of human intelligence and a significant step toward achieving general machine intelligence. Despite the remarkable success of modern neural networks across diverse domains, they often struggle when faced with unobserved data outside their training distribution. In this work, we investigate a general class of \textit{implicit} deep learning models \cite{NEURIPS2019_01386bd6,NEURIPS2018_69386f6b,chen2018neural,IDL} that encompasses classical layered-wise neural networks as special cases. Implicit deep learning models allow information to propagate both forwardly and backwardly through \textit{closed-form feedback loops}, offering a flexible and powerful approach to learning data representations.

Various formulations have been proposed, including deep equilibrium models (DEQs) \cite{NEURIPS2019_01386bd6}, Neural ODEs \cite{chen2018neural}, and implicit models \cite{IDL}. These models don't have explicitly defined layers and instead employ state vectors defined via an ``equilibrium'' (fixed-point) equation. In this framework, outputs are determined implicitly by the same equation. Recent results have highlighted the utility of implicit models \cite{NEURIPS2020_3812f9a5,10.5555/3495724.3496729,tsai2022state}. There has also been emerging work where the equilibrium state is interpreted as a closed-loop feedback system from a neuroscience perspective \cite{ma2022principles}. Moreover, emerging work interprets the equilibrium state as a closed-loop feedback system from a neuroscience perspective \cite{ma2022principles}. Unlike traditional feed-forward models that traverse nodes in a single direction in the computational graph, inputs of implicit models can revisit or reverse directions between nodes through closed-loop feedback.

In this paper, we investigate whether implicit deep learning models exhibit higher extrapolation capabilities, a fundamental skill in human intelligence, compared to similarly sized non-implicit neural network models. Our contributions are summarized in the following:
\begin{itemize}[noitemsep]
    \item We demonstrate the extrapolation power of implicit deep learning models across three distinct domains: well-defined mathematical arithmetic, real-world earthquake location data, and volatile time series forecasting.
    \item We conduct analyses and ablation studies on \textit{depth adaptability} and \textit{closed-loop feedback}, revealing that features learned by implicit models are more generalizable compared to their non-implicit counterparts.
\end{itemize}

\subsection{Related Work.}



\paragraph{Mathematical tasks.} Previous research has primarily focused on developing specialized neural network models capable of learning algorithms \cite{NTM,NTMSequel,NeuralGPU,ooddeq,Thinklonger}. For instance, Neural Arithmetic Logic Units (NALUs) were specifically designed to represent mathematical relationships within their architecture, with the goal of improving arithmetic extrapolation \cite{NALU}. However, these models were later found to be highly unstable during training \cite{iNALU}. Studies by \citeauthor{trans1} demonstrated that transformers performed effectively for addition and subtraction tasks, achieving high accuracy in interpolation experiments. However, challenges arose with other Transformer-based architectures such as BART \cite{trans2} and large language models (LLMs) \cite{wei2022emergent}, which struggled to accurately reproduce functions when dealing with inputs from a wide distribution range. On the contrary, \citeauthor{transeig} showed that Transformers could provide ``roughly correct'' solutions for matrix inversion and eigenvalue decomposition tasks, even on out-of-distribution (OOD) inputs, indicating a notable level of mathematical understanding.

\paragraph{Out-of-distribution generalization.} Only a handful of studies have begun to explore out-of-distribution (OOD) generalization for implicit deep learning models \cite{ooddeq,NeuralODE_OOD,NeuralODE_OOD2}. These works showcase the capabilities of implicit deep learning models on tasks like Blurry MNIST, sequential tasks \cite{ooddeq}, matrix inversion, and graph regression \cite{pi_deq}. Researchers such as \citeauthor{ooddeq} and \citeauthor{pi_deq} highlight a characteristic of DEQs known as path independence, where they converge to a similar fixed point regardless of initialization. This property suggests that if DEQs iterate for longer before converging on OOD inputs, they could gather more information about the data and potentially outperform other models. \cite{IDL_Peyre} theorized that this property exists only when testing DEQs on more complex data than in their training distribution. They demonstrate that increasing the number of inner iterations can lead to overfitting for interpolation tasks. We contribute further evidence supporting both hypotheses by examining well-defined functions and real-world data with out-of-distribution characteristics.

\paragraph{Function extrapolation.} \citeauthor{nn_extrap} studied how ReLU multi-layer perception (MLPs) and graph neural networks extrapolate on quadratic, cosine, and linear functions. They identified specific architectural choices that enhance extrapolation, such as encoding task-specific non-linearities in model features. Similarly, in the vision domain, \citeauthor{extrap_functions} introduced context normalization for more generalized features. Additionally, \citeauthor{polynets} demonstrated the advantages of neural networks with Hadamard products (NNs-Hp) and polynomial networks (PNNs) for arithmetic extrapolation. In this paper, we hypothesize that implicit models can ``adapt'' to distribution changes, thereby eliminating the need for specific feature transformations for extrapolation.

\subsection{Background}
Given a data input $u$, DEQs are defined as $f_w(u, x)$ where $f$ is a function that depends on model parameters $w$, and $x$ represents the hidden features. The objective is to find a fixed point $x^*$ by iteratively applying $f$ to $x$ until convergence, such that $x^*=f_w(u, x^*)$. In this paper, we adopt the more generalized formulation presented in \citeauthor{IDL}. This formulation, proven to converge linearly, includes DEQs as a special case. Formally, for a given data point $u \in \reals^p$, an implicit model solves the equilibrium equation $x = \phi(Ax + Bu)$, where $x$ represents the equilibrium state for input $u$, $\phi$ denotes a non-linear activation such as ReLU, and matrices $A$ and $B$ are model parameters. Prediction is obtained by passing the equilibrium state $x$ through an affine transformation, yielding $\hat{y}(u) \in \reals^q$, where matrices $C$ and $D$ are also model parameters:
\begin{equation}\label{eq:pred-rule}
\hat{y}(u) = Cx+Du , \mbox{ where } x = \phi(Ax+Bu).
\end{equation}
The ``state'' vector $x \in \reals^n$ contains the ``hidden features'' of the model. These features are not expressible in closed form but are implicitly defined via the equilibrium equation $x = \phi(Ax+Bu)$. 

In this paper, we define convergence as $\|x_t-x_{t-1}\|_{\infty} < \varepsilon$, where $x_t$ and $x_{t-1}$ represent the current and previous implicit hidden states, respectively. We set $\varepsilon = 3 \times 10^{-6}$ . As demonstrated by \citeauthor{IDL}, implicit models will always converge under certain conditions, such as projecting the $A$ matrix onto an infinite norm ball, which we ensure in our experiments. We train implicit models using stochastic gradient descent and back-propagation. During the forward pass, we aim to approximate the fixed point $x^*$ given a set of weights $A$, $B$, $C$, and $D$, along with a threshold $\varepsilon$. We denote $x_N$ as the approximation of the fixed point. During the backward pass, implicit differentiation is employed to compute the Jacobian of $x_N$ with respect to the model weights. We define $\Phi := \frac{\partial \phi(x_t)}{\partial x_t}$ as a block diagonal matrix of gradients with $t \in [0, N]$. As demonstrated by \citeauthor{IDL} in Appendix G, the gradient equations for each weight matrix are unique and rely on the inversion of $(I - \Phi A)^{-1}$, which can be accelerated through approximate inverse Jacobian methods \cite{fung2022jfb,jacob_2}. 
\section{Problem Setup}
\label{metho}

We explore two types of implicit models in this paper: standard implicit models as defined by Eq. (\ref{eq:pred-rule}), and a variant we term \textit{implicitRNN}.


\paragraph{ImplicitRNN.}
The \textit{implicitRNN} functions similarly to a vanilla RNN, processing sequence inputs one by one. For each time step $i$, where $s_i \in \reals^p$ represents the $i$-th element in a sequence $(s_1, s_2, \cdots, s_t)$. The model input $u$ for implicitRNN is a concatenation of $s_i$ and the previous hidden state $h_{i-1}$, akin to a vanilla RNN. The implicit prediction rule for implicitRNN is as follows:
\begin{align*}
    &h_0 = \mathbf{0} ; \quad x_0 = \mathbf{0} ; \quad u_i = \begin{pmatrix} s_i \\ h_{i-1} \end{pmatrix} \\
    &x = \phi(Ax + Bu_i) \quad \mbox{(equilibrium equation)} \\
    &\hat{y_i}(u_i) = Cx + Du_i \quad \mbox{(prediction equation)} \\
    &h_i = \hat{y_i}(u_i).
\end{align*}
In this setup, the recurrent layer is replaced with an implicit structure comprising the equilibrium and prediction equations. We introduce the implicitRNN to contrast implicit with explicit sequential models, both retaining representations of data timestep by timestep.

\subsection{Extrapolate on Mathematical Tasks}
We explore three types of functions, ranging from simple to complex: 1) identity function, 2) arithmetic operations, and 3) rolling functions over sequential data. To evaluate the extrapolation performance, we use two sets of data: $u_{\text{train}} \sim P(\mu; \sigma)$ and $u_{\text{test}} \sim P(\mu + \kappa; \sigma + \kappa)$, where $P$ represents a known distribution, $\mu$ and $\sigma$ are mean and the standard-deviation, and $\kappa$ denote the hyper-parameters for distribution shift. Our code, models, dataset, and experiment setup are available on GitHub\footnote{The link is redacted for anonymity. It will be released in our camera-ready version}.

\paragraph{Identify function.}
Recent work has shown that neural networks struggle to learn the basic task of identity mapping, $f(u) = u$, where models should return the exact input as given \cite{he2016identity,NALU}. 

\paragraph{Arithmetic operations}
We also focus on two arithmetic operations: \textit{addition} and \textit{subtraction} with additional data transformation. Following the task proposed by \citeauthor{NALU}, we randomly select four numbers $i, j, k, l$ from 1 to 50, ensuring that $i<j$ and $k<l$. For each sample, we construct two new numbers $a$ and $b$ from the input $\vec{u} := \langle u_1, u_2, \cdots, u_{50} \rangle$ as follows: $a = \sum_{a=i}^j u_a$, $b = \sum_{b=k}^l u_b$. Finally, we predict $y = a + b$ for addition and $y = a - b$ for subtraction. 

\paragraph{Rolling functions.}
Lastly, we explore two rolling functions over a sequence: \textit{average} and \textit{argmax}. In the rolling average task, we predict the average of the sequence up to the current time step $j$ for each timestep, calculated as $\sum_{i=1}^j u_i/j$. For the rolling argmax task, we predict the index of the maximum value seen by the model so far for each time step. We convert this into a classification problem by outputting a one-hot vector of length $L=10$, representing the index of the predicted maximum input seen so far, where $L$ is the length of the input sequence. Our evaluation focuses on predictions made for the final element of the sequence.

We compare implicit models with neural networks specially designed to excel on each task: MLPs for simple functions, LSTMs for sequential data \cite{LSTM}, NALU for OOD arithmetic task \cite{NALU}, and various Transformers-based models \cite{transformers}. The details of each task and the model architectures are provided in the Appendix. We optimize all models using grid search and 5-fold cross-validation on in-distribution inputs. To ensure a fair comparison, we set the model dimension of the implicit models to have a similar number of parameters as their non-implicit counterparts.


\subsection{Extrapolate on Noisy Real-world Data}
In addition to mathematical extrapolation where data is generated from a known underlying function, we consider extrapolation on real-world problems where data is typically noisy and often lacks known data generation functions. We focus on two real-world applications known for predicting unseen events: oscillating time series forecasting and earthquake location prediction

\begin{figure}[!h]
    \centering
    \includegraphics[width=.3\textwidth]{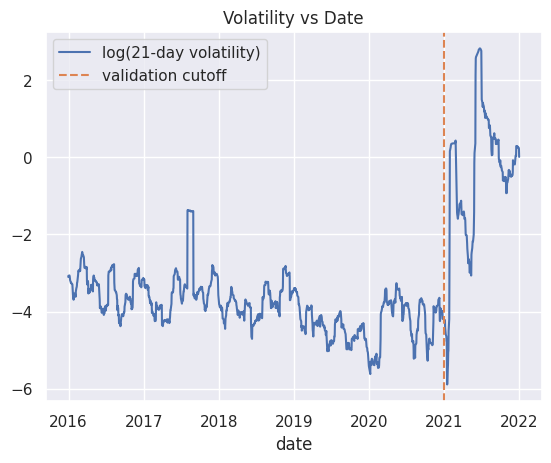}
    \caption{Time series of a 21-day rolling average of AMC stock volatility plotted on a log scale, highlights a drastic volatility increase at the beginning of our validation cutoff.}
    \label{fig:vol-plot}
\end{figure}

\paragraph{Oscillating time series forecasting.} We first consider a simpler synthetic version known as spiky time series forecasting, where spikes are inserted into a time series derived from a combination of sine functions at random intervals (see Appendix \ref{spiky-explained}). In the second case, we forecast AMC stock data volatility, particularly focusing on the drastic increase in average volatility observed at the beginning of 2021 as illustrated in Figure \ref{fig:vol-plot}.  Volatility, in this context, refers to the expected amount by which a security's price might change suddenly, serving as a measure of financial risk for an asset. Our prediction task involves forecasting AMC's volatility over the next 10 minutes based on its volume-weighted average trade price (VWAP) for each of the past 60 minutes. Volatility is calculated as the variance of the VWAP prices\footnote{Normally, volatility is calculated as the standard deviation of returns. However, we wish to amplify the changes in distribution between our train and test set, so we use raw prices and the variance instead.} during the forecast period. Our training data goes from 2/1/2015 to 12/31/2020 and our validation set from 1/1/2021 to 12/31/2021. Our training data spans from 2/1/2015 to 12/31/2020, with the validation set covering 1/1/2021 to 12/31/2021. We deliberately refrain from making our data stationary through differencing or examining returns, as we aim to evaluate our models' ability to adapt to changes in price distribution. We compare the performance of implicit models against simple linear regression and non-implicit neural network baselines, with details of the baseline models included in Table \ref{architectures} in the Appendix.

\paragraph{Earthquake location prediction.} The earthquake location prediction is a well-established problem in seismology \cite{Eiko,earthquake-pred1}. It involves predicting the location ($X$, $Y$, and $Z$ coordinates) and the seismic wave travel time ($T$) of an earthquake based on seismographs recorded from nearby seismometers. Successfully solving this problem can have significant humanitarian implications, allowing for early warnings before the potentially destructive second wave(s) of an earthquake. However, the problem remains challenging due to the sparsity of the event observations \cite{L_paper}. 


\begin{figure}[!h]
    \centering
    \includegraphics[width=.23\textwidth]{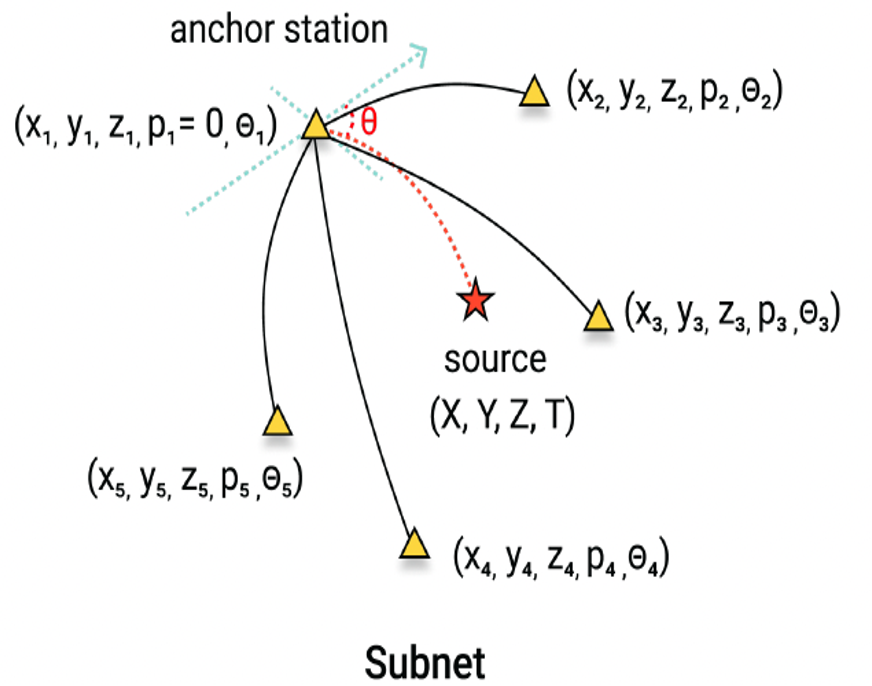}
    \includegraphics[width=.23\textwidth]{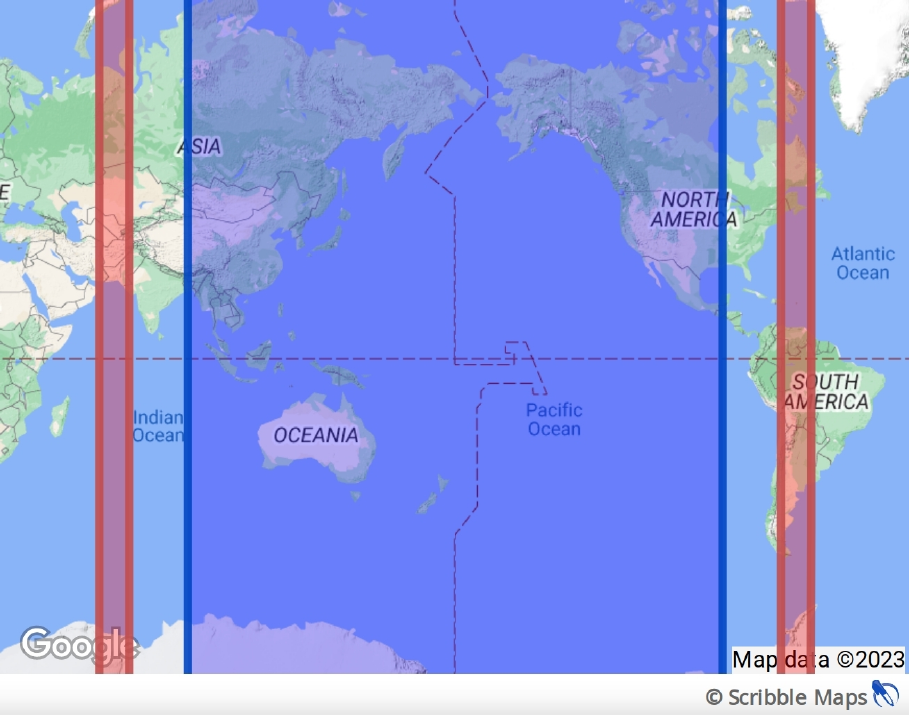}
    \caption{\textbf{Left}: Geometric visualization of one set of training features ($x_i$, $y_i$, $z_i$, $p_i$, $\theta_i$) and its corresponding labels ($X$, $Y$, $Z$, $T$). The triangles correspond to stations and the star corresponds to a source. \textbf{Right}: The map shows the training set region colored in blue, roughly corresponding to the Pacific Ring of Fire. The two red areas are the testing set regions for $k = 3$.}
    \label{fig:eqsetup}
\end{figure}

We follow the methodology outlined by \citeauthor{L_paper}, generating seismograph samples recorded by five stations with one anchor station designated as the reference for all seismic wave arrival times. As illustrated in the left panel of Figure \ref{fig:eqsetup}, our input features include a station's coordinates ($x$, $y$, $z$), event-station back-azimuths ($\theta$), and relative wave travel times ($p$) with respect to the anchor station. Training data is synthetically generated within a range of $90^{\circ}$E and $-90^{\circ}$E, roughly corresponding to the Pacific Ring of Fire, as shown in Figure \ref{fig:eqsetup} (further details are provided in Appendix \ref{earthquake-explained}). Testing is conducted on regions shifted from $10^{\circ}$E to $90^{\circ}$E beyond this Ring of Fire. As a comparison baseline, we use EikoNet, a deep learning model introduced by \citeauthor{Eiko} specifically designed for earthquake location prediction. While earthquakes primarily occur in certain active tectonic boundaries, an extrapolated earthquake location prediction system can aid in detecting earthquakes in new areas, whether natural or human-made (e.g., caused by mining, oil, and gas activities). Such a system is also valuable for explosion monitoring, providing universal mapping capabilities.

\section{Experimental Results}
\subsection{Mathematical Extrapolation}
\begin{figure}
    \centering
    \includegraphics[width=.3\textwidth]{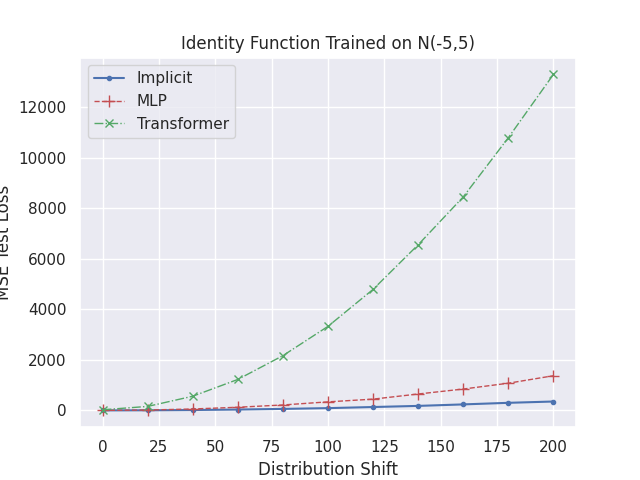}
    \caption{Test MSE for the identity function task. MSE for MLP and Transformers model increases as the distribution shift hyper-parameter $\kappa$ increases.}
    \label{fig:identiy-loss}
\end{figure}

\paragraph{Identify function.}
The test mean squared error (MSE) for the identify function task is shown in Figure \ref{fig:identiy-loss}. The implicit model maintains the lowest test MSE (below 5) for test data with a distribution shift from 0 to 25. Even for very large distribution shifts of up to 200, where $u_{\text{test}} \in \reals^{10} \sim U(-205, 205)$,  the implicit model outperforms the Transformers encoder model by a factor of $10^{5}$ and the MLP by a factor of $10^{3}$. This disparity highlights the implicit model's robustness in handling distribution shifts compared to non-implicit models, which tend to overfit to the training distribution and exhibit increased error with larger distribution shifts. The identity function task exposes the spurious features learned by both the MLP and Transformers models. Our implicit model reaches equilibrium after only 4 training iterations. In contrast, similar-sized MLPs and transformer encoders struggle with overfitting, highlighting the implicit model's ability to limit overfitting through faster convergence when dealing with simpler functions like identity mapping.

\begin{figure}
    \centering
    \includegraphics[width=.48\textwidth]{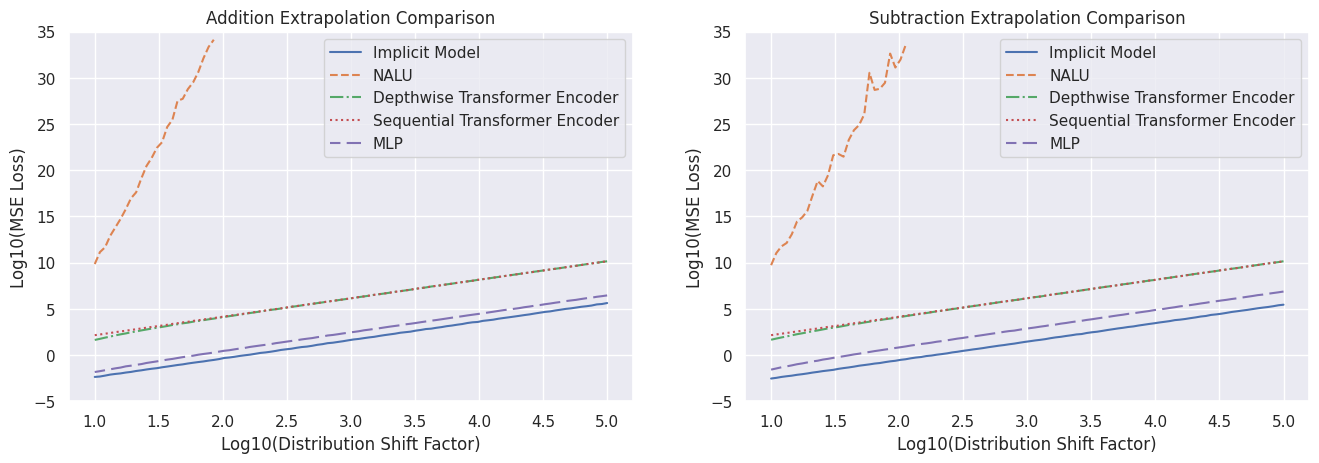}
    \caption{Test Log(MSE) for the arithmetic operations. The implicit model strongly outperforms all other models on OOD data.}
    \label{fig:arithmetic-extrap}
\end{figure}

\paragraph{Arithmetic operations.}
The results for the addition and subtraction arithmetic tasks are depicted in Figure \ref{fig:arithmetic-extrap}. The implicit model not only outperforms various Transformer encoders but also NALU, a model specifically designed for mathematical operations. As shown by \citeauthor{wei2022emergent}, Transformers models require much larger model sizes ($10^{23}$) to perform an arithmetic task effectively. In contrast, an implicit model with only $7,650$ parameters successfully learns these operations with a testing loss of $<1$ for shifts $<100$. Throughout the experiment, we consistently observed that implicit models outperform non-implicit models on extrapolation tasks, especially with limited training samples. Surprisingly, the specialized NALU model achieves the worst testing loss, exceeding $10^{10}$ for an extrapolation shift of only 10, as shown in Figure \ref{fig:arithmetic-extrap}. Across our experiments, we couldn't produce robust out-of-distribution predictions with the NALU model.

\begin{figure*}[!h]
    \begin{minipage}{0.64\linewidth}
        \centering
        \includegraphics[width=.49\textwidth]{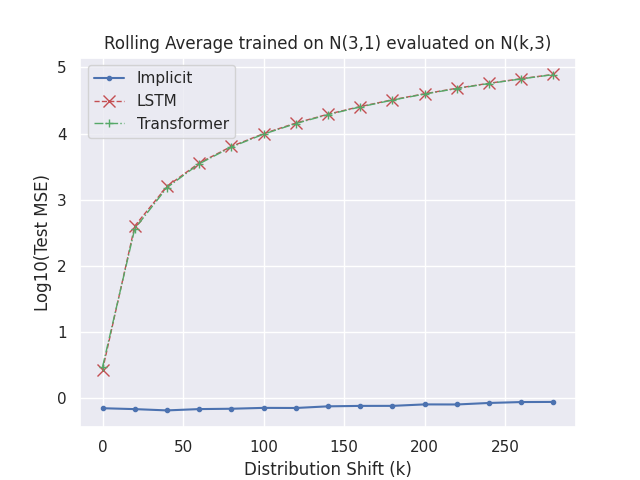}
        \includegraphics[width=.49\textwidth]{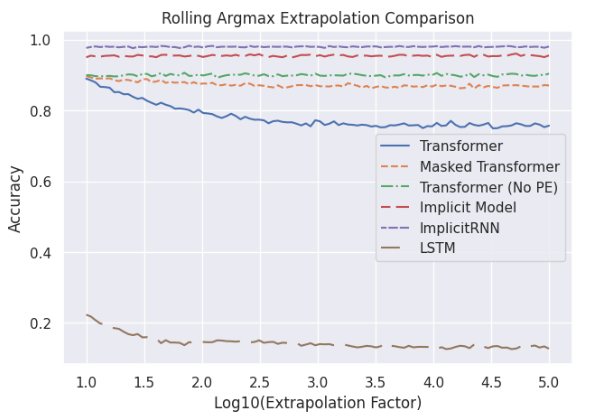}
        \caption{For rolling tasks, implicit models maintain close to constant loss ($\downarrow$) and accuracy ($\uparrow$) across shifts.}
        \label{fig:seq-plots}
    \end{minipage}%
    \hspace{.03\textwidth}%
    \begin{minipage}{0.32\linewidth}
        \includegraphics[width=.88\textwidth]
        {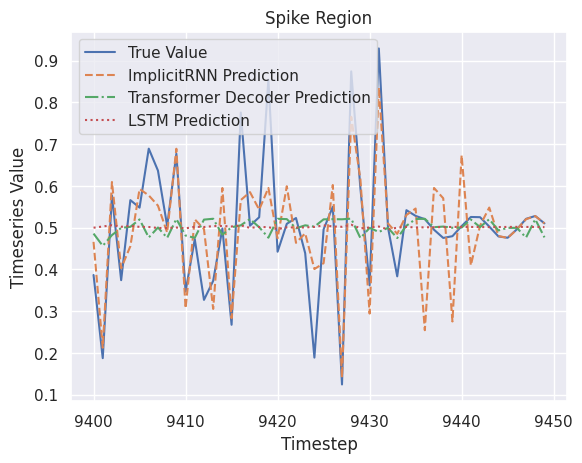}
        \caption{The implicitRNN most accurately models spike magnitudes.}
        \label{fig:spike-plots}
    \end{minipage}
\end{figure*}

\paragraph{Rolling functions.}
In Figure \ref{fig:seq-plots}, we present the performance of the two sequence modeling tasks. In the rolling average task (left plot of Figure \ref{fig:seq-plots}), the LSTM and Transformers exhibit similar behavior while the implicit model maintains close to constant loss regardless of the test distribution. For the rolling argmax, we use a Transformer encoder-decoder architecture, where the target sequence is the right-shifted argmax labels.  This architecture has an inherent advantage, as simply outputting the final element in this right-shifted sequence will provide the argmax of the input sequence up to but not including the current element. Given that argmax labels of a sequence are uniformly distributed, this setup implies an expected accuracy of 90\% or of $1 - 1/L$, where $L=10$ is the sequence length. Surprisingly, we observe that the standard implicit model and implicitRNN both outperform the LSTM and Transformer models. Additionally, the implicitRNN outperforms the standard implicit model, indicating the potential benefits of a rolling latent representation in tasks requiring awareness of relative positions within a sequence.

Comparing different Transformers variants highlights the potential for small Transformers to overfit to their positional encodings (PE) on simpler tasks. The Transformer without positional encodings, lacking the ability to differentiate between different positions in the sequence, outperforms the other two architectures, consistently hovering around the 90\% accuracy benchmark. This performance suggests that it effectively learns to always output the final value of its target sequence. Overall, these results demonstrate a very competitive performance of the standard implicit model and implicitRNN over Transformers and LSTMs in extrapolation tasks, particularly in scenarios where an understanding of relative positions within a sequence is crucial.

It is worth emphasizing that the implicitRNN structure resembles that of a vanilla RNN, with the standard recurrent cell replaced by an implicit layer. The substantial performance improvement over the LSTM suggests an exciting potential of the implicit layer to learn effective memory-gating mechanisms. However, further experimentation, specifically with longer sequence lengths, would be required to verify this conjecture. For further insights and experimental findings, please refer to Appendix \ref{more-results}.

\subsection{Oscillating Time Series Forecasting}

In our prior experiments, we tested implicit models on clearly defined mathematical operations. Now, we shift focus to real-world noisy data, where the underlying function is unknown. This transition allows us to explore whether the extrapolation benefits of implicit models extend to such complex, real-world scenarios.

In the spiky time series forecasting task, our objective is for models to effectively capture sudden and short-lived distribution changes in our generated data. Such extreme events, resembling those seen in stock prices, sales volumes, or storage capacity predictions, often challenge forecasting models that struggle to extrapolate to these unprecedented patterns. The ability to accurately adapt to synthetic spikes in this task carries significant real-life implications. Table \ref{spiky-table} demonstrates that implicitRNN achieves a threefold reduction in test MSE compared to non-implicit models in the spiky time series task. Furthermore, Figure \ref{fig:spike-plots} showcases the implicit model's capacity to accurately capture the locations and magnitudes of these spikes. In contrast, the transformer decoder and LSTM models often output the average of the time series. This tendency of standard models to revert to the mean also hints at their potential weakness in simulating time-series data effectively. Encouraged by these positive results, we applied implicit models to predict the sharp rise in AMC stock volatility observed in 2021. We report the mean absolute percentage error (MAPE)\footnote{The Mean Absolute Percentage Error (MAPE) metric assesses a model's capability to predict changes in magnitudes relative to their sizes. Consequently, it is not directly comparable across distributions. Despite the higher Root Mean Square Error (RMSE) on the validation data (for the implicit model: train RMSE =  0.001784, validation RMSE = 0.266366), the validation MAPE is lower since the average volatility in the validation set is orders of magnitude larger.} in Table \ref{spiky-table}, where the implicit model outperforms other baselines by a factor of 1.67.

\begin{table}[!h]
   \centering
   \resizebox{.35\textwidth}{!}{
    \begin{tabulary}{\textwidth}{LRRRR}
        \toprule
         & \multicolumn{2}{c}{\textbf{Spiky Data}} & \multicolumn{2}{c}{\textbf{AMC Stock}} \\
         & \multicolumn{2}{c}{(MSE)} & \multicolumn{2}{c}{(MAPE)} \\
        \cmidrule(lr){2-3} \cmidrule(lr){4-5}
         \textbf{Model} & Train & Test & Train & Test \\
        \midrule
        Transformer &  0.061 &  0.012 & 12.2 & 6.51\\
        ImplicitRNN & 0.015 & \textbf{0.004} & \textbf{2.61} & \textbf{1.71} \\
        LSTM & \textbf{0.011} & 0.011 & 10.5 & 5.46 \\
        MLP & - & - & 5.51 & 2.87 \\
        Linear regression & - & - & 7.19 & 3.94 \\
        \bottomrule         
    \end{tabulary}
    }
    \caption{\label{spiky-table} Train and test metrics ($\downarrow$) in forecasting time series with sudden changes for synthetic (spiky time series) and real-world data (AMC stock volatility). Our architectures vary between tasks (see Table \ref{architectures}) but the implicit model outperforms all fine-tuned models.}
\end{table}

\subsubsection{Earthquake Location Prediction}
\begin{figure}[!h]
    \centering
    \includegraphics[width=.23\textwidth]{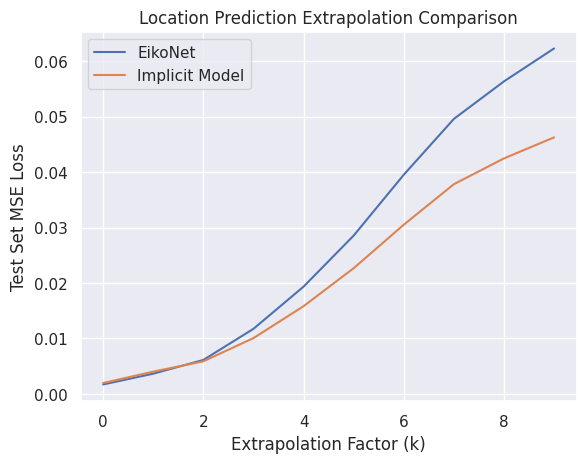}
    \includegraphics[width=.23\textwidth]{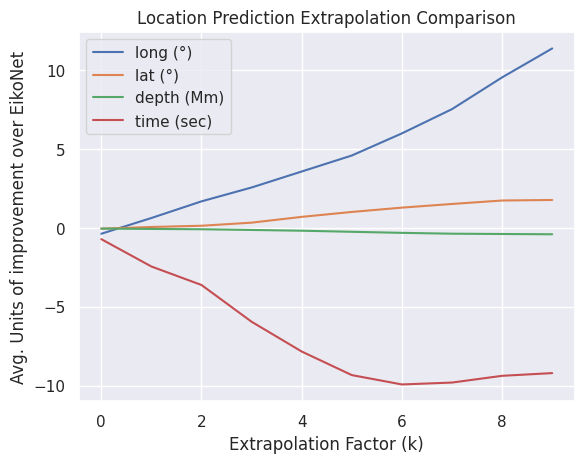}
    \caption{Extrapolation comparison between EikoNet and implicit model on the location prediction task as the extrapolation factor increases. \textbf{Left}: The implicit model has the general edge in terms of MSE loss. \textbf{Right}: Breaking down the prediction values, we observe that the implicit model only outperforms EikoNet in longitude and latitude predictions.}
    \label{fig:eqresults}
\end{figure}

In the earthquake location prediction task, our implicit model demonstrates an improvement of $0.25\mathrm{e}{-3}$ in the in-distribution test loss compared to EikoNet, a model specifically designed for seismic data. Notably, in the left panel of Figure \ref{fig:eqresults}, as $k$ increases, we observe a progressive improvement in the performance of the implicit model in terms of the extrapolated test MSE. By the time $k$ reaches 2, the implicit model has surpassed EikoNet, and at $k$ = 9, the implicit model's test loss is $1.59\mathrm{e}{-2}$ better than that of EikoNet. This improvement translates to an average enhancement of $11^{\circ}$ in longitude and $2^{\circ}$ in latitude. Further refinement could be achieved by limiting the source latitude along with longitude during training, potentially leading to a more substantial improvement in latitude extrapolation. However, as depicted in the right panel of Figure \ref{fig:eqresults}, the implicit model encounters greater difficulty in predicting time and depth, exhibiting deteriorating performance with increasing $k$. Specifically, at $k$ = 9, the implicit model's average performance worsens by 9.2 seconds and 409 km. Future research should explore whether training an implicit model exclusively on time and depth labels can enhance its performance in these aspects. This two-pass approach resembles traditional location prediction software such as HYPOINVERSE (\citeauthor{latlon}), where constraints on depth and time are notably challenging.

\section{Analysis}
\def\mcirc{\mathbin{\scalerel*{\circ}{j}}}

In our analysis, we pinpoint two key properties that contribute to the effective extrapolation capabilities of implicit models, even with limited datasets. The first property is "depth adaptability," meaning these models are not constrained by a fixed number of layers. The second property, ``closed-loop feedback'', allows inputs to revisit the same node within a single pass through the model, enhancing its learning and adaptability.

\begin{figure*}[!h]
    \centering
    \includegraphics[width=.32\textwidth]{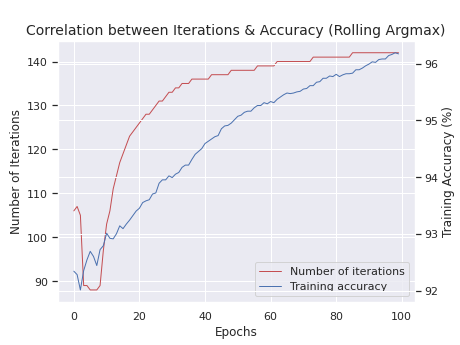}
    \vspace{.2cm}
    \includegraphics[width=.33\textwidth]{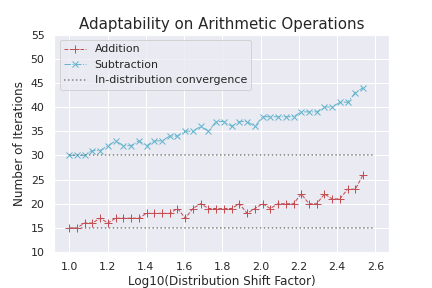}
    \vspace{-.2cm}
    \includegraphics[width=.33\textwidth]{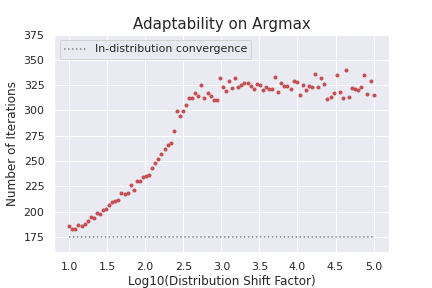}
    \caption{\textbf{Left}: Relationship between model performance during training and the number of iterations. \textbf{Middle} and \textbf{right}: Growth of implicit models as the input complexity increases in the arithmetic tasks and rolling argmax.}
    \label{fig:adapt}
\end{figure*}

\paragraph{Depth adaptability.} During a forward pass through an implicit model, the process concludes either upon convergence to a fixed point $x^*$ or upon reaching a predetermined iteration limit. We investigated whether the complexity of the input affects the number of iterations required for the model to achieve equilibrium. Across our experiments, we noticed that as our model familiarized itself with parameter matrices, the iteration count stabilized for inputs within the distribution. On average, implicit models converged in about 15 iterations for learning the addition operation, 30 iterations for subtraction, and 175 iterations for the rolling argmax task. We can interpret the number of iterations required for a given task as an indication of the model's perceived difficulty with that task. Consistent with the findings of \citeauthor{ooddeq}, we observed that as our input deviated further from the training distribution, the number of iterations needed to converge increased. This suggests that the input $U$ underwent more transformations by the model parameters (as shown in the bottom two panels of Figure \ref{fig:adapt}). Therefore, we interpret that implicit models dynamically ``grow'' in depth to adapt their feature space to extrapolated inputs, iterating longer compared to in-distribution inputs. This adaptability allows them to benefit from low depth for in-distribution inputs (reducing overfitting) and higher depth for out-of-distribution inputs (reducing underfitting).

\begin{figure}[!h]
    \centering
    \includegraphics[width=.3\textwidth]{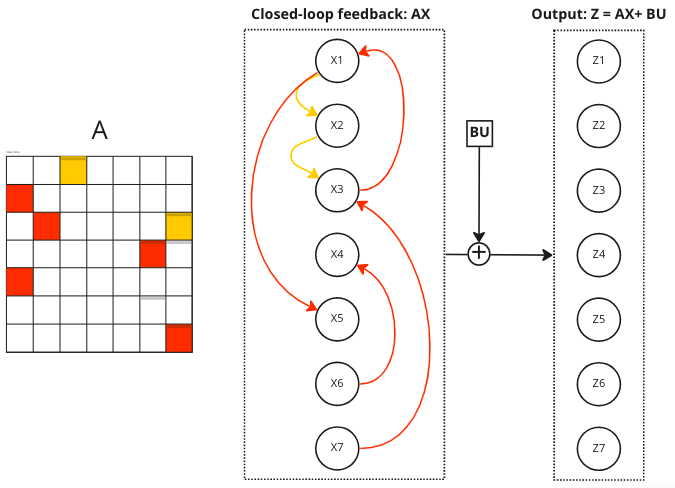}
    \caption{A matrix and node diagram representation of one iteration through the implicit model: $Ax+Bu$. In red, we have the weights that induce feedback in $A$, and in yellow, the non-feedback weights.}
    \label{fig:implicit-model-graphs}
\end{figure}

\paragraph{Closed-loop feedback.} Introduced by \citeauthor{cybernetics}, refers to a system's ability to self-correct based on its outputs. This concept, elaborated by \citeauthor{cybernetics2}, involves returning a portion of the output, which partly influences the input at the next step. In neural networks, feedback is encoded through the backward pass. Implicit models, as depicted in Figure \ref{fig:implicit-model-graphs}, inherently incorporate feedback within a single iteration via the feedback connection in the parameter matrix $A$ (lower-triangular part). We observe that inputs in implicit models do not only move forward from one set of nodes to the next, as in a standard feed-forward neural network. Instead, they may revisit the same nodes or even move backward, allowing the model to correct itself from one layer to the next within a single iteration. This is in contrast to non-implicit models, which correct themselves after a pass through all layers.  \citeauthor{ma2022principles} highlighted closed-loop feedback as a strength of implicit models and noted its similarity to human neural networks. Therefore, we evaluated the utility of closed-loop feedback in extrapolation through ablation studies on the lower triangular part of the model's $A$ matrix. We illustrate the absence of feedback in an implicit model by considering a $3\times3$ example with a strictly upper triangular $A$ matrix. At time $t+1$, the iteration equation simplifies to $x^{t+1} = \phi(Ax^{t} + Bu)$. Here, we can disregard the activation $\phi$ and $Bu$ terms, focusing solely on the $x^{t}$ term as it carries the outputs from the previous iteration. This setup allows us to demonstrate how an implicit model without feedback operates, contrasting it with the typical behavior of a truly implicit models.

\small
\begin{align}
\label{eq:feedback}
    x^{t+1} \approx Ax^{t} &= 
\begin{pmatrix}
    x^{t+1}_0 \\
    x^{t+1}_1 \\
    x^{t+1}_2 \\
\end{pmatrix}
=
\begin{pmatrix}
    \star & W_{0} & W_{1}\\ 
    \star & \star & W_{2} \\
    \star & \star & \star\\
\end{pmatrix} 
\begin{pmatrix}
    x^{t}_0 \\
    x^{t}_1 \\
    x^{t}_2 \\
\end{pmatrix} \\
&=
\begin{pmatrix}
    \star \cdot x^{t}_0 + W_0 x^{t}_1 + W_1 x^{t}_2\\
    \star \cdot x^{t}_0 + \star \cdot x^{t}_1 + W_2 x^{t}_2 \\
    \star \cdot x^{t}_0 + \star \cdot x^{t}_1 + \star \cdot x^{t}_2 \nonumber \\
\end{pmatrix}.
\end{align}
\normalsize
Closed-loop feedback corresponds to $x_i^{t}$ being used to generate $x_i^{t+1}$. Equation \ref{eq:feedback} illustrates how the $\star$ weights encode this feedback mechanism. Specifically, the $i$-th state at time step $t+1$, denoted as $x_i^{t+1}$, depends not only on its past output $x_i^{t}$ directly through weights on the diagonal but also indirectly through weights on the lower triangular matrix. As an example of this indirect dependence, consider $x_2^{t+1}$, which depends not only on $x_2^{t}$ but also on $x_1^{t}$ due to the lower triangular $\star$ weights. Furthermore, both $x_1^{t}$ and $x_2^{t}$ depend on $x_0^{t-1}$, highlighting the chain of dependencies facilitated by the feedback mechanism.

\begin{figure}[!h]
    \hspace{-1.8em}
    \includegraphics[width=.55\textwidth]{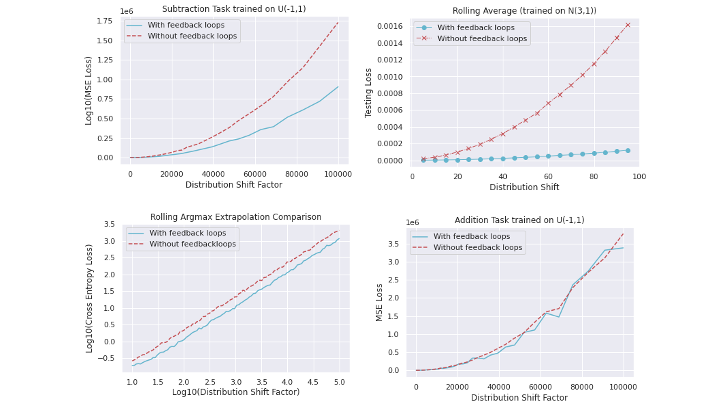}
    \footnotesize
    \caption{Implicit models benefit from closed-loop feedback specifically in harder extrapolation tasks.}
    \label{fig:arithmetic-feedback}
\end{figure}

We compared a standard implicit model (with feedback loops) against an ablated implicit model (without feedback loops) where the upper-triangularity of $A$ was enforced during training. The standard implicit models correspond to those used in our experiments, as described in Section \ref{metho}. The results of the mathematical tasks for implicit models with and without feedback are presented in Table \ref{ablation-study}. It is evident that the feedback loops play a crucial role in helping the models achieve significantly lower testing loss, especially on inputs with distribution shifts. Figure \ref{fig:arithmetic-feedback} presents the ablation results for both models across distribution shifts in arithmetic and rolling sequential tasks. Ablating the feedback harms model performance for subtraction but not for addition. The regular model, with its feedback loops, has twice as many weights, which could potentially lead to overfitting on simpler tasks. Feedback loops appear to increase stability to distribution shifts for rolling average tasks, whereas they only provide better performance in calculating the rolling argmax of a sequence. We also note that besides superior overall performance, the presence of closed-loop feedback seems to make the model more resistant to distribution shifts: notably, we observe that its loss increases slower in the subtraction and rolling average experiments. Our analysis demonstrates that implicit models possess the capability to adapt their architecture by learning the necessary depth, node connections, and correcting themselves based on past predictions within a single training iteration, thanks to their closed-loop feedback mechanism.

\begin{table}[!h]
    \centering
    \resizebox{.3\textwidth}{!}{
    \begin{tabulary}{\textwidth}{LCC}
        \toprule
        \textbf{Task} & \textbf{Feedback} & \textbf{No feedback}\\
        \midrule
        Add. & 3.06 & 4.07 \\
        Sub. & 0.953 & 1.80 \\
        R. Max. & 93.7\% & 91.8\% \\
        R. Avg. & \num{1e-4} &  \num{1e-3} \\
        \bottomrule
    \end{tabulary}
    }
    \caption{\label{ablation-study} Testing metrics (MSE $\downarrow$ for the arithmetic operations and rolling average, accuracy $\uparrow$ for rolling argmax) for a distribution shift of 100 comparing an implicit model trained with and without closed-loop feedback.}
\end{table}

Implicit models represent a significant advancement over standard feed-forward neural network layers by relying on the convergence of an equilibrium equation to extract features from input data. In this paper, we have demonstrated their remarkable ability to learn useful function representations for extrapolation tasks, even in scenarios with limited training data. Our experiments consistently show that implicit models outperform non-implicit baselines when handling out-of-distribution inputs, temporal and geographical shifts. The adaptive nature of training implicit models allows them to explore and identify optimal architectures with minimal hand-engineering, resulting in a robust inner function representation for extrapolation tasks involving out-of-distribution data. These results highlight the potential of implicit models and motivate further research into their use as more interpretable solutions for wider scenarios with limited data set.


\clearpage
\newpage
\bibliographystyle{named}
\bibliography{references}

\begin{thebibliography}{}

\bibitem[\protect\citeauthoryear{Anil \bgroup \em et al.\egroup }{2022}]{pi_deq}
Cem Anil, Ashwini Pokle, Kaiqu Liang, Johannes Treutlein, Yuhuai Wu, Shaojie Bai, Zico Kolter, and Roger Grosse.
\newblock Path independent equilibrium models can better exploit test-time computation.
\newblock In {\em Advances in Neural Information Processing Systems (NeurIPS 2022)}, 2022.

\bibitem[\protect\citeauthoryear{Bai \bgroup \em et al.\egroup }{2019}]{NEURIPS2019_01386bd6}
Shaojie Bai, J.~Zico Kolter, and Vladlen Koltun.
\newblock Deep equilibrium models.
\newblock In H.~Wallach, H.~Larochelle, A.~Beygelzimer, F.~d\textquotesingle Alch\'{e}-Buc, E.~Fox, and R.~Garnett, editors, {\em Advances in Neural Information Processing Systems}, volume~32. Curran Associates, Inc., 2019.

\bibitem[\protect\citeauthoryear{Bai \bgroup \em et al.\egroup }{2020}]{NEURIPS2020_3812f9a5}
Shaojie Bai, Vladlen Koltun, and J.~Zico Kolter.
\newblock Multiscale deep equilibrium models.
\newblock In H.~Larochelle, M.~Ranzato, R.~Hadsell, M.F. Balcan, and H.~Lin, editors, {\em Advances in Neural Information Processing Systems}, volume~33, pages 5238--5250. Curran Associates, Inc., 2020.

\bibitem[\protect\citeauthoryear{Charton}{2022}]{transeig}
François Charton.
\newblock What is my math transformer doing? -- three results on interpretability and generalization, 2022.

\bibitem[\protect\citeauthoryear{Chen \bgroup \em et al.\egroup }{2018a}]{NEURIPS2018_69386f6b}
Ricky T.~Q. Chen, Yulia Rubanova, Jesse Bettencourt, and David~K Duvenaud.
\newblock Neural ordinary differential equations.
\newblock In S.~Bengio, H.~Wallach, H.~Larochelle, K.~Grauman, N.~Cesa-Bianchi, and R.~Garnett, editors, {\em Advances in Neural Information Processing Systems}, volume~31. Curran Associates, Inc., 2018.

\bibitem[\protect\citeauthoryear{Chen \bgroup \em et al.\egroup }{2018b}]{chen2018neural}
Ricky~TQ Chen, Yulia Rubanova, Jesse Bettencourt, and David~K Duvenaud.
\newblock Neural ordinary differential equations.
\newblock {\em Advances in neural information processing systems}, 31, 2018.

\bibitem[\protect\citeauthoryear{Chuang \bgroup \em et al.\egroup }{2023}]{L_paper}
Lindsay~Y. Chuang, Jesse Williams, Louisa Barama, Zhigang Peng, and Andrew~V. Newman.
\newblock Deep learning and masksembles for sparse network earthquake location and uncertainty estimation.
\newblock Manuscript in preparation., 2023.

\bibitem[\protect\citeauthoryear{El~Ghaoui \bgroup \em et al.\egroup }{2021}]{IDL}
Laurent El~Ghaoui, Fangda Gu, Bertrand Travacca, Armin Askari, and Alicia Tsai.
\newblock Implicit deep learning.
\newblock {\em SIAM Journal on Mathematics of Data Science}, 3(3):930--958, 2021.

\bibitem[\protect\citeauthoryear{Fung \bgroup \em et al.\egroup }{2022}]{fung2022jfb}
Samy~Wu Fung, Howard Heaton, Qiuwei Li, Daniel McKenzie, Stanley Osher, and Wotao Yin.
\newblock Jfb: Jacobian-free backpropagation for implicit networks.
\newblock In {\em Proceedings of the AAAI Conference on Artificial Intelligence}, volume~36, pages 6648--6656, 2022.

\bibitem[\protect\citeauthoryear{Geng \bgroup \em et al.\egroup }{2021}]{jacob_2}
Zhengyang Geng, Xin-Yu Zhang, Shaojie Bai, Yisen Wang, and Zhouchen Lin.
\newblock On training implicit models.
\newblock In M.~Ranzato, A.~Beygelzimer, Y.~Dauphin, P.S. Liang, and J.~Wortman Vaughan, editors, {\em Advances in Neural Information Processing Systems}, volume~34, pages 24247--24260. Curran Associates, Inc., 2021.

\bibitem[\protect\citeauthoryear{Graves \bgroup \em et al.\egroup }{2014}]{NTM}
Alex Graves, Greg Wayne, and Ivo Danihelka.
\newblock Neural turing machines.
\newblock {\em CoRR}, abs/1410.5401, 2014.

\bibitem[\protect\citeauthoryear{Graves \bgroup \em et al.\egroup }{2016 10}]{NTMSequel}
Alex Graves, Greg Wayne, Malcolm Reynolds, Tim Harley, Ivo Danihelka, Agnieszka Grabska-Barwińska, Sergio~Gómez Colmenarejo, Edward Grefenstette, Tiago Ramalho, John Agapiou, Adrià~Puigdomènech Badia, Karl~Moritz Hermann, Yori Zwols, Georg Ostrovski, Adam Cain, Helen King, Christopher Summerfield, Phil Blunsom, Koray Kavukcuoglu, and Demis Hassabis.
\newblock Hybrid computing using a neural network with dynamic external memory.
\newblock {\em Nature.}, 538(7626), 2016-10.

\bibitem[\protect\citeauthoryear{Gu \bgroup \em et al.\egroup }{2020}]{10.5555/3495724.3496729}
Fangda Gu, Heng Chang, Wenwu Zhu, Somayeh Sojoudi, and Laurent El~Ghaoui.
\newblock Implicit graph neural networks.
\newblock In {\em Proceedings of the 34th International Conference on Neural Information Processing Systems}, NIPS'20, Red Hook, NY, USA, 2020. Curran Associates Inc.

\bibitem[\protect\citeauthoryear{He \bgroup \em et al.\egroup }{2016}]{he2016identity}
Kaiming He, Xiangyu Zhang, Shaoqing Ren, and Jian Sun.
\newblock Identity mappings in deep residual networks.
\newblock In {\em European conference on computer vision}, pages 630--645. Springer, 2016.

\bibitem[\protect\citeauthoryear{Hochreiter and Schmidhuber}{1996}]{LSTM}
Sepp Hochreiter and J\"{u}rgen Schmidhuber.
\newblock Lstm can solve hard long time lag problems.
\newblock In M.C. Mozer, M.~Jordan, and T.~Petsche, editors, {\em Advances in Neural Information Processing Systems}, volume~9. MIT Press, 1996.

\bibitem[\protect\citeauthoryear{Kaiser and Sutskever}{2015}]{NeuralGPU}
Łukasz Kaiser and Ilya Sutskever.
\newblock Neural gpus learn algorithms, 2015.

\bibitem[\protect\citeauthoryear{Kelly \bgroup \em et al.\egroup }{2020}]{NeuralODE_OOD2}
Jacob Kelly, Jesse Bettencourt, Matthew~James Johnson, and David Duvenaud.
\newblock Learning differential equations that are easy to solve, 2020.

\bibitem[\protect\citeauthoryear{Kennett \bgroup \em et al.\egroup }{1995}]{earthquake_paper}
B.~L.~N. Kennett, E.~R. Engdahl, and R.~Buland.
\newblock {Constraints on seismic velocities in the Earth from traveltimes}.
\newblock {\em Geophysical Journal International}, 122(1):108--124, 07 1995.

\bibitem[\protect\citeauthoryear{Liang \bgroup \em et al.\egroup }{2021}]{ooddeq}
Kaiqu Liang, Cem Anil, Yuhuai Wu, and Roger Grosse.
\newblock Out-of-distribution generalization with deep equilibrium models.
\newblock {\em ICML 2021 Workshop on Uncertainty and Robustness in Deep Learning}, 2021.

\bibitem[\protect\citeauthoryear{Ma \bgroup \em et al.\egroup }{2022}]{ma2022principles}
Yi~Ma, Doris Tsao, and Heung-Yeung Shum.
\newblock On the principles of parsimony and self-consistency for the emergence of intelligence.
\newblock {\em Frontiers of Information Technology \& Electronic Engineering}, pages 1--26, 2022.

\bibitem[\protect\citeauthoryear{Nogueira \bgroup \em et al.\egroup }{2021}]{trans1}
Rodrigo Nogueira, Zhiying Jiang, and Jimmy Lin.
\newblock Investigating the limitations of the transformers with simple arithmetic tasks.
\newblock {\em CoRR}, abs/2102.13019, 2021.

\bibitem[\protect\citeauthoryear{Patten and Odum}{1981}]{cybernetics2}
B.~Patten and E.~Odum.
\newblock The cybernetic nature of ecosystems.
\newblock 118(6), December 1981.

\bibitem[\protect\citeauthoryear{Rackauckas \bgroup \em et al.\egroup }{2021}]{NeuralODE_OOD}
Christopher Rackauckas, Yingbo Ma, Julius Martensen, Collin Warner, Kirill Zubov, Rohit Supekar, Dominic Skinner, Ali Ramadhan, and Alan Edelman.
\newblock Universal differential equations for scientific machine learning, 2021.

\bibitem[\protect\citeauthoryear{Ramzi \bgroup \em et al.\egroup }{2023}]{IDL_Peyre}
Zaccharie Ramzi, Pierre Ablin, Gabriel Peyré, and Thomas Moreau.
\newblock Test like you train in implicit deep learning.
\newblock 2023.

\bibitem[\protect\citeauthoryear{Saad \bgroup \em et al.\egroup }{2021}]{earthquake-pred1}
Omar~M. Saad, Ali~G. Hafez, and M.~Sami Soliman.
\newblock Deep learning approach for earthquake parameters classification in earthquake early warning system.
\newblock {\em IEEE Geoscience and Remote Sensing Letters}, 18(7):1293--1297, 2021.

\bibitem[\protect\citeauthoryear{Schlör \bgroup \em et al.\egroup }{2020}]{iNALU}
Daniel Schlör, Markus Ring, and Andreas Hotho.
\newblock inalu: Improved neural arithmetic logic unit.
\newblock {\em Frontiers in Artificial Intelligence}, 3, 2020.

\bibitem[\protect\citeauthoryear{Schwarzschild \bgroup \em et al.\egroup }{2021}]{Thinklonger}
Avi Schwarzschild, Eitan Borgnia, Arjun Gupta, Furong Huang, Uzi Vishkin, Micah Goldblum, and Tom Goldstein.
\newblock Can you learn an algorithm? generalizing from easy to hard problems with recurrent networks.
\newblock In M.~Ranzato, A.~Beygelzimer, Y.~Dauphin, P.S. Liang, and J.~Wortman Vaughan, editors, {\em Advances in Neural Information Processing Systems}, volume~34, pages 6695--6706. Curran Associates, Inc., 2021.

\bibitem[\protect\citeauthoryear{Smith \bgroup \em et al.\egroup }{2021}]{Eiko}
Jonathan~D. Smith, Kamyar Azizzadenesheli, and Zachary~E. Ross.
\newblock {EikoNet}: Solving the eikonal equation with deep neural networks.
\newblock {\em {IEEE} Transactions on Geoscience and Remote Sensing}, 59(12):10685--10696, dec 2021.

\bibitem[\protect\citeauthoryear{Trask \bgroup \em et al.\egroup }{2018}]{NALU}
Andrew Trask, Felix Hill, Scott~E Reed, Jack Rae, Chris Dyer, and Phil Blunsom.
\newblock Neural arithmetic logic units.
\newblock In S.~Bengio, H.~Wallach, H.~Larochelle, K.~Grauman, N.~Cesa-Bianchi, and R.~Garnett, editors, {\em Advances in Neural Information Processing Systems}, volume~31. Curran Associates, Inc., 2018.

\bibitem[\protect\citeauthoryear{Tsai \bgroup \em et al.\egroup }{2022}]{tsai2022state}
Alicia~Y Tsai, Juliette Decugis, Laurent~El Ghaoui, and Alper Atamt{\"u}rk.
\newblock State-driven implicit modeling for sparsity and robustness in neural networks.
\newblock {\em arXiv preprint arXiv:2209.09389}, 2022.

\bibitem[\protect\citeauthoryear{{USGS}}{}]{latlon}
{USGS}.
\newblock Hypoinverse earthquake location.

\bibitem[\protect\citeauthoryear{Vaswani \bgroup \em et al.\egroup }{2017}]{transformers}
Ashish Vaswani, Noam Shazeer, Niki Parmar, Jakob Uszkoreit, Llion Jones, Aidan~N Gomez, \L~ukasz Kaiser, and Illia Polosukhin.
\newblock Attention is all you need.
\newblock In I.~Guyon, U.~Von Luxburg, S.~Bengio, H.~Wallach, R.~Fergus, S.~Vishwanathan, and R.~Garnett, editors, {\em Advances in Neural Information Processing Systems}, volume~30, pages 5998--6008. Curran Associates, Inc., 2017.

\bibitem[\protect\citeauthoryear{Wang \bgroup \em et al.\egroup }{2021}]{trans2}
Cunxiang Wang, Boyuan Zheng, Yuchen Niu, and Yue Zhang.
\newblock Exploring generalization ability of pretrained language models on arithmetic and logical reasoning.
\newblock In Lu~Wang, Yansong Feng, Yu~Hong, and Ruifang He, editors, {\em Natural Language Processing and Chinese Computing}, pages 758--769, Cham, 2021. Springer International Publishing.

\bibitem[\protect\citeauthoryear{Webb \bgroup \em et al.\egroup }{2020}]{extrap_functions}
Taylor~W. Webb, Zachary Dulberg, Steven~M. Frankland, Alexander~A. Petrov, Randall~C. O'Reilly, and Jonathan~D. Cohen.
\newblock Learning representations that support extrapolation.
\newblock {\em CoRR}, abs/2007.05059, 2020.

\bibitem[\protect\citeauthoryear{Wei \bgroup \em et al.\egroup }{2022}]{wei2022emergent}
Jason Wei, Yi~Tay, Rishi Bommasani, Colin Raffel, Barret Zoph, Sebastian Borgeaud, Dani Yogatama, Maarten Bosma, Denny Zhou, Donald Metzler, Ed~H. Chi, Tatsunori Hashimoto, Oriol Vinyals, Percy Liang, Jeff Dean, and William Fedus.
\newblock Emergent abilities of large language models.
\newblock {\em Transactions on Machine Learning Research}, 2022.
\newblock Survey Certification.

\bibitem[\protect\citeauthoryear{Wiener}{1948}]{cybernetics}
N.~Wiener.
\newblock Cybernetics.
\newblock {\em MIT Press, Cambridge, Mass}, 1948.

\bibitem[\protect\citeauthoryear{Wu \bgroup \em et al.\egroup }{2022}]{polynets}
Yongtao Wu, Zhenyu Zhu, Fanghui Liu, Grigorios Chrysos, and Volkan Cevher.
\newblock Extrapolation and spectral bias of neural nets with hadamard product: a polynomial net study.
\newblock In Alice~H. Oh, Alekh Agarwal, Danielle Belgrave, and Kyunghyun Cho, editors, {\em Advances in Neural Information Processing Systems}, 2022.

\bibitem[\protect\citeauthoryear{Xu \bgroup \em et al.\egroup }{2021}]{nn_extrap}
Keyulu Xu, Mozhi Zhang, Jingling Li, Simon~S. Du, Ken ichi Kawarabayashi, and Stefanie Jegelka.
\newblock How neural networks extrapolate: From feedforward to graph neural networks.
\newblock 2021.

\end{thebibliography}

\newpage
\appendix
\section{Appendix}



\newcommand{\tabitem}{~~\llap{\textbullet}~~}

\subsection{Model Specificites}

We used ReLU activation function for our implicit models, transformers have a dropout of 0.1 and a layer norm epsilon of $\epsilon = 10^{-5}$. See Tables \ref{distributions} and \ref{architectures} on the next page for model architecture and experiment specificities.

\begin{table*}[!h]
    \renewcommand{\arraystretch}{1.2}
    \centering
     \begin{tabularx}{\textwidth}{mbb}
    \toprule
    \textbf{Task} &\textbf{Training Distribution} &\textbf{Testing Distribution} \\
    \midrule
    Identity Function
    & $u_{\text{train}}\in\reals^{10,000 \times 10}\sim U(-5, 5)$
    & $u_{\text{test}} \in \reals^{3,000 \times 10} \sim U(-\kappa, \kappa)$, where $\kappa$ ranges from 10 to 80 \\ 
    Arithmetic Operations
    & $u_{\text{train}} \in \reals^{10,000 \times 50} \sim U(-1, 1)$
    & $u_{\text{test}} \in \reals^{3,000 \times 50} \sim U(-\kappa/2, \kappa/2)$, $\kappa$ ranges from $10$ to $10^5$ \\
    Rolling Average 
    & $u_{\text{train}} \in \reals^{10,000 \times 10} \sim \mathcal{N}(3,1)$ 
    & $u_{\text{test}} \in \reals^{3,000 \times 10} \sim \mathcal{N}(3+\kappa,1)$, $\kappa$ ranges from 5 to 100 \\
    Rolling Argmax 
    & $u_{\text{train}} \in \reals^{10,000 \times 10} \sim U(0,1)$ 
    & $u_{\text{test}} \in \reals^{3,000 \times 10} \sim U(0,\kappa)$, $\kappa$ ranges from $10^1$ to $10^5$ \\    
    Earthquake Location 
    & 720,576 ($X$, $Y$, $Z$) locations sampled between (90, -90)$^\circ$E, 30 features
    & 20,016 samples in each extrapolation region $(90 - 10\kappa,  100 - 10\kappa) \cup (-100 + 10\kappa, -90 + 10k)$, $\kappa$ ranges from 1 to 9 \\
    \bottomrule
    \end{tabularx}
    \caption{\label{distributions} Details of training and out-of-distribution test set for each extrapolation task. }
\end{table*}

\begin{table*}[!h]
    \renewcommand{\arraystretch}{1.2}
    \centering
     \begin{tabularx}{\textwidth}{smbb}
    \toprule
    \textbf{Task} &\textbf{Baseline model} &\textbf{Implicit models} &\textbf{Transformers} \\

    \cmidrule{1-4}
    Identity Function & MLP: $10 \times 9 \times 9 \times 10$ & Regular: $A \in \reals^{4 \times 4}, B \in \reals^{4 \times 10}, C \in \reals^{10 \times 4}, D \in \reals^{10 \times 10}$ & Encoder-decoder: $10 \times 10 \times 5$, 5 attention heads 
    \\\cmidrule{1-4}
    
    Arithmetic Operations & 
    \tabitem MLP: $50 \times 10 \times 10 \times 1$ 
    & Regular: $A \in \reals^{20 \times 20}, B \in \reals^{20 \times 50}, C \in \reals^{1 \times 20}, D \in \reals^{1 \times 50}$ 
    & \tabitem Sequential encoder: 1 layer, 10 attention heads, feedforward dim 50 - processes each array as a single sequence \\
    
    & \tabitem NALU: $50 \times 10 \times 10 \times 1$ & & \tabitem Depth-wise encoder: 1 layer, 1 attention head, feedforward dim 500, max PE length 50 - processes each element in a given array as a single sequence \\
    
    \\\cmidrule{1-4}
    
    Rolling Average & LSTM: 1$\times$18$\times$18$\times$1 & Regular: $A \in \reals^{32 \times 32}, B \in \reals^{32 \times 10}, C \in \reals^{10 \times 32}, D \in \reals^{10 \times 10}$ & Encoder-decoder: $10 \times 10 \times 5 \times 10$, $5$ attention heads 
    
    \\\cmidrule{1-4}
    
    Rolling Argmax & LSTM: 1$\times$21$\times$21$\times$10 & \tabitem Regular: $A \in \reals^{36 \times 36}, B \in \reals^{36 \times 10}, C \in \reals^{10 \times 36}, D \in \reals^{10 \times 10}$ & \tabitem Masked encoder-decoder: 1 encoder layer, 1 decoder layer, 2 attention heads, feedforward dim 10, max PE length 10 \\

    & & \tabitem RNN: $A \in \reals^{21 \times 21}, B \in \reals^{21 \times 23}, C \in \reals^{22 \times 21}, D \in \reals^{22 \times 23}$ & \tabitem Unmasked encoder-decoder: 1 encoder layer, 1 decoder layer, 2 attention heads, feedforward dim 10, max PE length 10 \\ & & & \tabitem Unmasked encoder-decoder without PE: 1 encoder layer, 1 decoder layer, 2 attention heads, feedforward dim 10
    
    \\\midrule
    
    Spiky Time Series & LSTM: 1$\times$20$\times$20$\times$1 & RNN: $A \in \reals^{20 \times 20}, B \in \reals^{20 \times 21}, C \in \reals^{20 \times 20}, D \in \reals^{20 \times 21}$ with a 20$\times$1 linear layer & 1x10 linear layer (expansion) $\xrightarrow[]{}$ masked decoder (1 layer, 2 attention heads, feedforward dim 2048, max PE length 10) $\xrightarrow[]{}$ 10x1 linear layer (contraction) \\

    \\\midrule
    
    Volatility Prediction
    & \tabitem LSTM: $1 \times 38 \times 38 \times 1$
    & \tabitem Regular: $A \in \reals^{53 \times 53}, B \in \reals^{53 \times 60}, C \in \reals^{1 \times 53}, D \in \reals^{1 \times 60}$
    & Sequential encoder (1 layer, 1 attention head, feedforward dim 2048, max PE length 60) $\xrightarrow[]{}$ 60x1 linear layer \\

    & \tabitem SGD Linear Regression & \tabitem RNN: $A \in \reals^{37 \times 37}, B \in \reals^{37 \times 41}, C \in \reals^{40 \times 37}, D \in \reals^{40 \times 41}$ with a 40$\times$1 linear layer
    & \\

    & \tabitem MLP: $60 \times 50 \times 27 \times 27 \times 27 \times 10 \times 1$ \\ 

    \\\midrule

    Earthquake Location Prediction 
    &  EikoNet: $270 \times 32 \times 128 \times 128 \times 128 \times 32 \times 4$ (42,500)
    &  Regular: $A \in \reals^{190 \times 190}, B \in \reals^{190 \times 270}, C \in \reals^{4 \times 190}, D \in \reals^{4 \times 270}$ (42,680) \\
    
    \bottomrule
    \end{tabularx}
    \caption{\label{architectures} Details of the explicit and implicit network architectures used in our experiments. }
\end{table*}

\subsection{Additional Experimental Results}
\label{more-results}

\begin{figure}[!h]
    \centering
    \includegraphics[width=.33\textwidth]{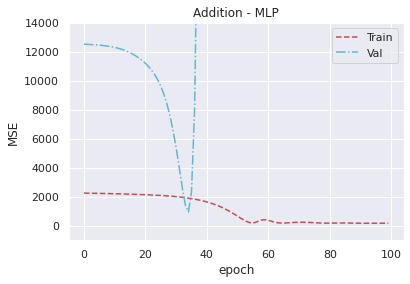}
    \includegraphics[width=.33\textwidth]{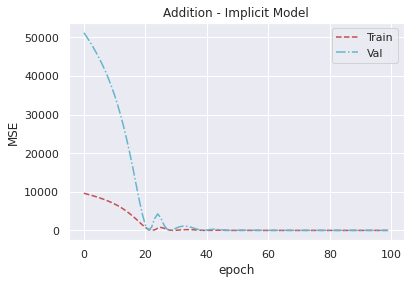}
    \includegraphics[width=0.33\textwidth]{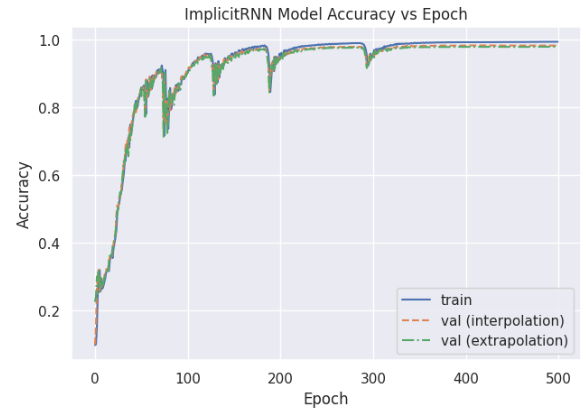}
    \caption{\textbf{Top} and \textbf{middle}: The MLP test loss explodes whereas the implicit model achieves testing loss close to 0. \textbf{Bottom}: For rolling argmax prediction with extrapolation factor t = 10, our implicitRNN performs similarly on interpolated and extrapolated data.}
    \label{fig:addition}
\end{figure}

We provide more in-depth results on the OOD generalization capacities of implicit models for a specific small distribution shift in Figure \ref{fig:addition}. We compare the training and validation loss on the addition task of both implicit and MLP models where $u_{\text{train}} \in \reals^{100} \sim U(1, 2)$ and $u_{\text{val}} \in \reals^{100} \sim U(2, 5)$. Even with this small distribution shift, we observe a large improvement.

\subsection{Spiky Data Generation}
\label{spiky-explained}

Both the LSTM and the implicit model were trained on 7000 data points and tested on 3000 data points. The training regime featured 20 spiky regions of 100 data points each. The testing regime featured a proportionate amount of spiky regions. The data points in the spiky regions were sampled from $y = 5 \times ( \sin(2x) + \sin(23x) + \sin(78x) + \sin(100x))$. We arbitrarily choose frequencies in $[0, 100]$ to generate a sufficiently spiky pattern. The magnitude of the spiky regions is at most 20. For the non-spiky regimes, the data points were sampled from $y = \sin(x)$ with added noise $\epsilon \sim \mathcal{N}(0, 0.25)$.

\subsection{Earthquake Data Generation}
\label{earthquake-explained}

To generate samples of seismic waves between specific longitudes, based on the methods presented by \citeauthor{L_paper}, we used a 1D velocity model called Ak135 from the Python library obspy.taup. Obspy is a Python framework used to process seismological data. \citeauthor{earthquake_paper} demonstrate the accuracy of this model compared to real-world data (see specifically Figure 6). A 1D velocity model assumes the P-wave travel time (the duration the P-wave takes to travel from point A to point B) only depends on two attributes: the distance between the source and the receiver station and the depth of the source. We use this model to create a travel time lookup table based on these two attributes. We then generate source locations from a mesh that spans the entire globe while adding perturbation to each latitude and longitude pair. We generate station locations using the source-station distances we have from the lookup table and place the stations in random orientations (azimuths) from the source.




\end{document}